\documentclass[10pt,twocolumn,letterpaper]{article}

\usepackage{iccv}
\usepackage{times}
\usepackage{epsfig}
\usepackage{graphicx}
\usepackage{amsmath}
\usepackage{amssymb}
\usepackage{CJKutf8} % 写中文
\usepackage{multirow} 
\usepackage{array}
\usepackage{color}
\definecolor{MyCyan}{RGB}{224,255,255}
\definecolor{MyLightRed}{rgb}{1,0.88,0.95}
\definecolor{MyGray}{RGB}{240,240,240}
\usepackage{colortbl}
\usepackage{xcolor}
\usepackage{booktabs}
\usepackage{bm}
\usepackage{bbding}
\usepackage{subfigure}
\usepackage{float}
\usepackage{amsthm,amsmath,amssymb}
\usepackage{mathrsfs}
\usepackage[algo2e,linesnumbered,ruled,noend]{algorithm2e}
\usepackage{bbm}
\usepackage[pagebackref=true,breaklinks=true,letterpaper=true,colorlinks,bookmarks=false]{hyperref}
\usepackage[accsupp]{axessibility}  % Improves PDF readability for those with disabilities.
\iccvfinalcopy % *** Uncomment this line for the final submission

\ificcvfinal\fi

\begin{document}

%%%%%%%%% TITLE
\title{GPA-3D: Geometry-aware Prototype Alignment for Unsupervised Domain Adaptive 3D Object Detection from Point Clouds}

\author{
	Ziyu Li$^{1}$\thanks{Work was jointly done when Ziyu was an intern at Huawei Noah's Ark Lab}\quad
	Jingming Guo$^{2}$\quad
	Tongtong Cao$^{2}$\quad
	Liu Bingbing$^{2}$\quad
	Wankou Yang$^{1}$\thanks{Corresponding Author}\quad
	\\
	$^{1}$ School of Automation, Southeast University \quad
	$^{2}$ Huawei Noah's Ark Lab
	\\
	{\tt\small \{liziyu, wkyang\}@seu.edu.cn} \quad
	\\
	{\tt\small \{guojingming, caotongtong, liu.bingbing\}@huawei.com}
}

\maketitle
\ificcvfinal\thispagestyle{empty}\fi

%%%%%%%%% ABSTRACT
\begin{abstract}
   	LiDAR-based 3D detection has made great progress in recent years. 
   	However, the performance of 3D detectors is considerably limited when deployed in unseen environments, owing to the severe domain gap problem.
   	Existing domain adaptive 3D detection methods do not adequately consider the problem of the distributional discrepancy in feature space, thereby hindering generalization of detectors across domains.
   	In this work, we propose a novel unsupervised domain adaptive \textbf{3D} detection framework, namely \textbf{G}eometry-aware \textbf{P}rototype \textbf{A}lignment (\textbf{GPA-3D}), which explicitly leverages the intrinsic geometric relationship from point cloud objects to reduce the feature discrepancy, thus facilitating cross-domain transferring.
	Specifically, GPA-3D assigns a series of tailored and learnable prototypes to point cloud objects with distinct geometric structures.
	Each prototype aligns BEV (bird's-eye-view) features derived from corresponding point cloud objects on source and target domains, reducing the distributional discrepancy and achieving better adaptation.
   	The evaluation results obtained on various benchmarks, including Waymo, nuScenes and KITTI, demonstrate the superiority of our GPA-3D over the state-of-the-art approaches for different adaptation scenarios.
   	The MindSpore version code will be publicly available at \url{https://github.com/Liz66666/GPA3D}.
\end{abstract}

%%%%%%%%% BODY TEXT
\section{Introduction}

As a fundamental research in 3D scene understanding, 3D detection from point clouds has attracted increasing attention due to its essential role in intelligent robotics, augmented reality and autonomous driving~\cite{survey_guo2020deep,survey_mao20223d,survey_lidarpointclouds,survey_3dpointclouds,survey_arnold2019survey}.
Despite significant process, state-of-the-art 3D detectors still suffer from dramatic performance degradation when training data and test data are from different environments, \textit{i.e.}, domain shift problem~\cite{uda3det_sn}.
Various factors, such as diverse weather conditions, object sizes, laser beams, and scanning patterns, lead to substantial discrepancies across different domains, hindering the transferability of existing LiDAR-based 3D detectors.
Intuitively, fine-tuning the detectors with adequate data from the target domain could alleviate this issue. 
However, manually annotating a large amount of point cloud scenes is a prohibitively expensive task.
Therefore, the research on unsupervised domain adaptation (UDA) for LiDAR-based 3D detection is essential.
\begin{figure}[t]	
	\centering
	\includegraphics[width=\linewidth]{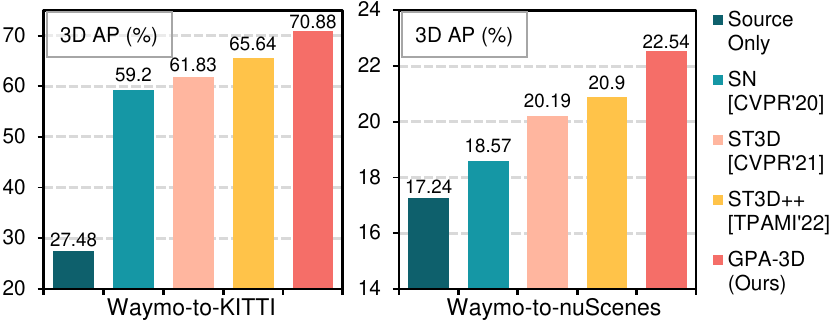}
	\vspace{-5mm}	
	\caption{The performance comparison with previous works~\cite{uda3det_sn,uda3det_st3d,3duda_st3d++}. 
		The detection architecture is SECOND-IoU~\cite{second,uda3det_st3d}.
		%
%		Please refer to Tab.~\ref{tab_waymo-to-kitti_comparison} and Tab.~\ref{tab_waymo-to-nuscenes_comparison} for more detailed comparisons.
		}\label{fig_main_results}
	\vspace{-6mm}
\end{figure}

Although many works have been proposed to deal with the UDA for image-based detection~\cite{cyclegan,huang2018auggan,rodriguez2019domain,li2022stepwise,hsu2020progressive,kim2019self,khodabandeh2019robust,strong_weak,dafasterrcnn}, directly applying these methods to 3D point cloud detection is insufficient for tackling the domain shifts.
These approaches mainly concentrate on the gaps of lighting and texture variations, which could not be obtained from point clouds.
While there is only a limited number of literature~\cite{uda3det_sn,uda3det_st3d,uda3det_sfuda3d,uda3det_uncertainty,3duda_3dcoco,3duda_st3d++,3duda_mlcnet} dealing with the UDA on LiDAR-based 3D detection. 
Prior work~\cite{uda3det_sn} utilizes the statistics from target annotation to perform data-level normalization.
MLC-Net~\cite{3duda_mlcnet} designs a mean-teacher framework to provide reliable pseudo-labels to facilitate transferring.
ST3D~\cite{uda3det_st3d} and ST3D++~\cite{3duda_st3d++} propose a self-training pipeline with a memory bank to collect and refine pseudo-labels.
Despite their great success, these methods do not adequately consider the problem of distributional discrepancy in feature space, hampering the adaptation performance.

\begin{figure}[t]	
	\centering
	\includegraphics[width=\linewidth]{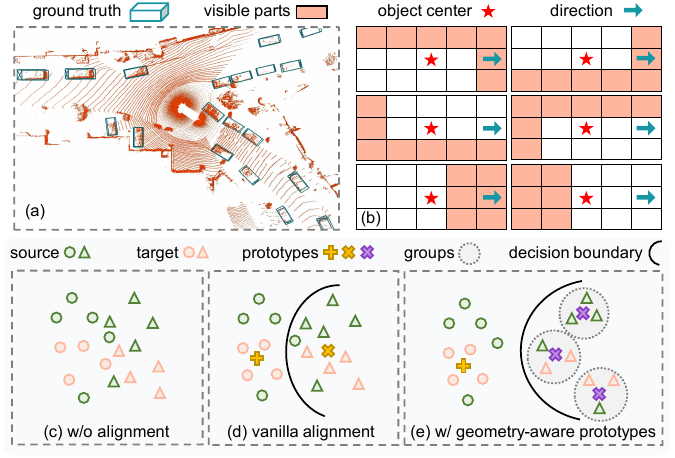}
	\vspace{-5mm}	 
	\caption{(a) Point cloud scene on BEV (bird's-eye-view).
	(b) Distinct geometric structures of point cloud objects.
	(c-e) Illustration of the distributional discrepancy.
	With explicit geometric constraints, the features from different domains are better aligned. }\label{fig_point_cloud_distribution}
	\vspace{-5mm}
\end{figure}

To reduce this discrepancy in 2D UDA task, some approaches utilize the class-wise prototypes align features from different domains~\cite{2d_uda_seg_proca,2d_uda_prototype_oriented,2d_uda_detection_mttrans,2d_uda_prototype_continual}.
In these works, a universal prototype is employed to enforce high representational similarity among features belonging to the same category.
However, in the case of 3D scenes, such as vehicles on the road, diverse locations and directions can result in distinct geometric structures, \textit{i.e.}, distributional patterns of point clouds, as presented in Fig.~\ref{fig_point_cloud_distribution} (a) and (b).
If a uniform prototype is applied to objects with completely different geometric structures, the efficacy of feature alignment might be hindered, as illustrated in Fig.~\ref{fig_point_cloud_distribution} (c-d).
We argue that adopting different prototypes to point cloud objects with distinct geometric structures could deal with the problem of distributional discrepancy, but more attention should also be paid to model these geometric structures during adaptation.

Based on the considerations, we propose a novel UDA framework for LiDAR-based 3D detectors, namely Geometry-aware Prototype Alignment (GPA-3D). 
Concretely, we first explore the potential relationships between the geometric structures of point cloud objects.
During training, we randomly extract the BEV features of point clouds from both the source and target domains,  and subsequently divide them into distinct groups based on their geometric structures.
In this process, BEV features derived from point clouds with the similar geometric structures will be classified into the same group. 
Each group is then assigned a unique prototype, which enforces high representational similarity among the BEV features within that group, as illustrated in Fig.~\ref{fig_point_cloud_distribution} (e).
To this end, the soft contrast loss is devised to pull the intra-group feature-prototype pairs closer in the representational space and push the inter-group pairs farther away.
Additionally, we develop the framework with two components, namely noise sample suppression (NSS) and instance replacement augmentation (IRA).
NSS utilizes the similarities between foreground areas and the background prototype, to produce a mask for decreasing the impact of noise.
IRA displaces pseudo-labels with high-quality samples that have similar geometric structures, enriching the diversity on the target domain.

The main contributions of this paper include:
\begin{itemize} 	
	\item We propose a novel UDA framework for LiDAR-based 3D detectors, namely Geometry-aware Prototype Alignment (GPA-3D).
	It explicitly integrates geometric associations into feature alignment, effectively decreasing the distributional discrepancy and facilitating the adaptation of existing point cloud detectors.
	 
	\item Noise sample suppression and instance replacement augmentation are designed to enhance pseudo-labels in terms of reliability and versatility, respectively.
	
	\item We conduct comprehensive experiments on Waymo, nuScenes, and KITTI. 
	The encouraging results demonstrate the GPA-3D outperforms state-of-the-art methods in various adaptation scenarios. 
	More importantly, thanks to the architecture-agnostic design, GPA-3D is flexible to be applied to point cloud detectors.
\end{itemize}

\begin{figure*}[t]	
	\centering
	\includegraphics[width=\textwidth]{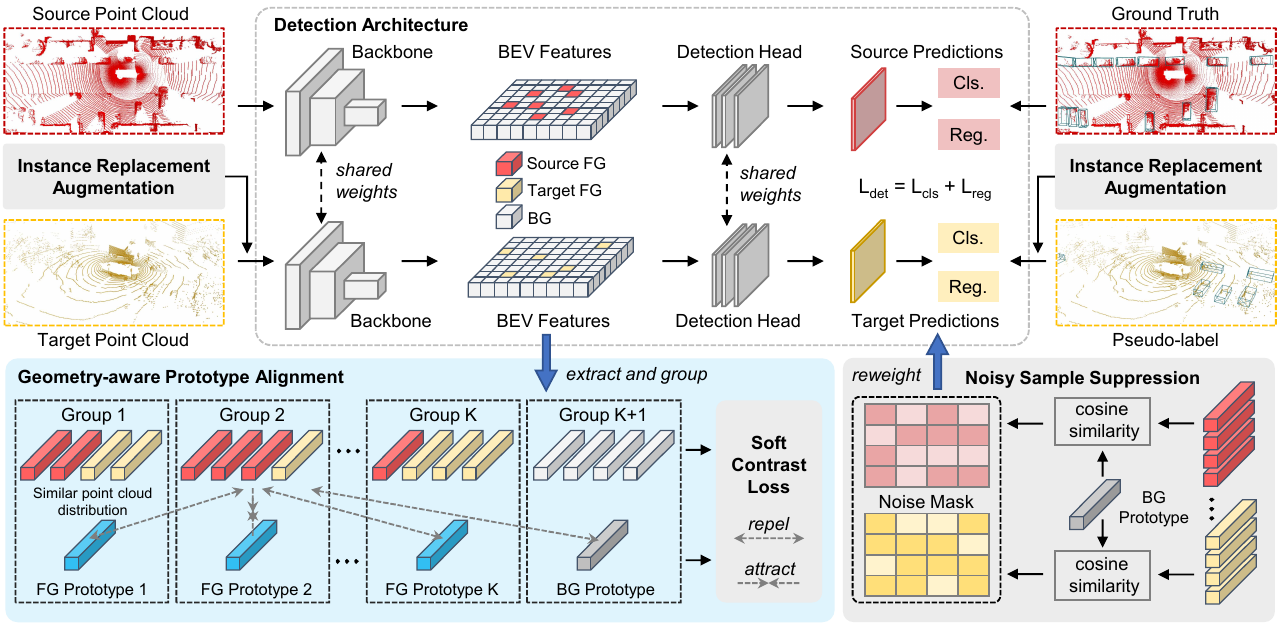}
	\vspace{-4mm}
	\caption{Overview of our proposed GPA-3D framework.
	It adopts a basic co-training manner to adapt the 3D detector with the point clouds from source and target domains.
	The BEV features are processed via geometry-aware prototype alignment, which reduces the distributional discrepancy and enables the learning of general representation across domains.
	To this end, the soft contrast loss is devised for jointly optimizing the prototypes and network parameters.
	Besides, the noisy sample suppression is proposed to alleviate the impact of noisy samples during training, and instance replacement augmentation is designed to enhance the diversity on the target domain.
	}\label{fig_framework}
	\vspace{-4mm}
\end{figure*}

\section{Related Work}
\paragraph{LiDAR-based 3D Detection.}
Mainstream point cloud detectors can be broadly divided into two categories: point-based and grid-based.
Point-based methods mainly adopt the architectures of PointNet~\cite{pointnet} and PointNet++~\cite{pointnet++} to extract features from raw point clouds.
PointRCNN~\cite{pointrcnn} designs an encoder-decoder backbone to learn the point-wise representation.
3DSSD~\cite{3dssd} improves the point sampling operator from the aspect of feature distance.
IA-SSD~\cite{3ddet_iassd} utilizes the instance-aware downsampling to preserve more foregrounds. 
On the other hand, grid-based methods first divide point clouds into fixed-size voxels, which are then processed via 2D/3D CNN.
SECOND~\cite{second} adopts sparse 3D convolution for efficient feature learning.
PointPillars~\cite{pointpillars} proposes a pillar encoding method and achieves a good trade-off between speed and performance.
PV-RCNN~\cite{pvrcnn} incorporates the voxel backbone with the keypoint branch to learn the representative scene features.
Other approaches~\cite{3ddet_pixor,3ddet_rangedet,3ddet_rsn} project the point clouds into certain kinds of 2D views, and employ 2D CNN to extract the features.
In this work, we conduct focused discussions with
SECOND~\cite{second} and PointPillars~\cite{pointpillars} as base detectors.
To demonstrate the generalization ability of our method, we also provide the comparisons with PV-RCNN~\cite{pvrcnn} detector.

\paragraph{Domain Adaptive Object Detection.} 
A large amount of literature has been presented in UDA for 2D-image detection, which can be roughly classified into two groups: distribution alignment and self-training.
Alignment-based methods~\cite{dafasterrcnn,strong_weak} leverage the adversarial training~\cite{domainadversarial} to learn aligned features across domains.
Self-training approaches~\cite{khodabandeh2019robust,kim2019self} utilize a multi-phase strategy to generate pseudo-labels on unlabeled data.
Besides, some works~\cite{hsu2020progressive,li2022stepwise,rodriguez2019domain,huang2018auggan} adopt the CycleGAN~\cite{cyclegan} to generate training samples with styles of source and target domains.
Similarly, several recent works also aim to address the domain bias for 3D point cloud detectors.
Wang \textit{et al.}~\cite{uda3det_sn} investigate the
domain bias of popular autonomous driving 3D datasets, and propose to alleviate the gaps via three techniques, \textit{i.e.}, output transformation, statistical normalization and few shot. 
SF-UDA$ ^{3D} $~\cite{uda3det_sfuda3d} adopts a mature 3D tracker to find the best scaling parameter, which is further used to re-scale the target point clouds for producing high-quality pseudo-labels.
MLC-Net~\cite{3duda_mlcnet} designs a mean-teacher paradigm to provide pseudo-labels for facilitating smooth learning of the student model.
ST3D~\cite{uda3det_st3d} and ST3D++~\cite{3duda_st3d++} build a self-training pipeline to produce pseudo-labels for fine-tuning model and update pseudo-labels via memory bank.
3D-CoCo~\cite{3duda_3dcoco} devises the domain-specific encoders with a hard sample mining strategy to learn transferable representations.
Compared with previous works, our method explicitly embraces the geometric relationship to reduce the distributional discrepancy during adaptation.

\section{The Proposed Method}
\vspace{-2.5mm}
In the following, we present GPA-3D to mitigate the domain gap for LiDAR-based detectors.
Fig.~\ref{fig_framework} illustrates the whole pipeline.
Sec.~\ref{sec:problem_statement} formulates the UDA task for point cloud detectors.
Sec.~\ref{sec:architecture} introduces the detection architecture in our method.
In Sec.~\ref{sec:prototype}, we explain the details of the geometry-aware prototype alignment, followed by the soft contrast loss, which is discussed in Sec.~\ref{sec:loss}.
Finally, we present the noise sample suppression and instance replacement augmentation in Sec.~\ref{sec:suppression} and Sec.~\ref{sec:aug}, respectively.

\subsection{Problem Statement}
\vspace{-2mm}
\label{sec:problem_statement}
In this work, we focus on the problem of unsupervised domain adaptation on 3D detection.
Concretely, given the labeled source domain point clouds $ \mathbb{D}^s = \{(P_i^s, L_i^s)\}_{i=1}^{N^s} $, as well as unlabeled target domain point clouds $ \mathbb{D}^t = \{(P_i^t)\}_{i=1}^{N^t} $, our goal is to train a 3D detector based on $ \mathbb{D}^s$ and $ \mathbb{D}^t$ and maximize its performance on $ \mathbb{D}^t$.
Here, $ N $ is the total number of scenes, and $ P_i $ indicates the $ i $-th point cloud scene, where each point has the 3-dim spatial coordinates and an extra intensity.
The corresponding label $ L_i $ represents a series of 3D bounding boxes, each of them can be parameterized by the center location ($ c_x, c_y, c_z $), spatial dimension ($ h, w, l $) and rotation $ r $.
Note that the superscripts $ s $ and $ t $ stand for source and target domain respectively.

\subsection{Detection Architecture}

\label{sec:architecture}
The input point cloud $ P_i $ is first sent to a backbone network with 3D sparse convolutions or 2D convolutions to extract the point cloud representation as following:
\vspace{-3mm}
\begin{equation}
	\bm{F}_i = h_1(P_i;\theta_1),
	\vspace{-2mm}
\end{equation}
where $ h_1 $ is the backbone with parameters $ \theta_1 $, and $ \bm{F}_i $ indicates the BEV features.
After that, a detection head $ h_2 $ with parameters $ \theta_2 $ produces the final output, formulated as:
\vspace{-2mm}
\begin{equation}
	\{b, s\}_i = h_2(\bm{F}_i;\theta_2),
	\vspace{-2mm}
\end{equation}
where $ b $ and $ s $ represent the predicted 3D boxes and scores respectively.
A co-training paradigm is applied to progressively mitigate the domain shift.
In each mini-batch, both the source point clouds $ P_i^s $ and target point clouds $ P_i^t $ are sent to the detector, and their outputs are supervised by the corresponding ground truth and pseudo-labels, respectively.

\subsection{Geometry-aware Prototype Alignment}
\label{sec:prototype}

\paragraph{Extract.} 
As mentioned in Sec.~\ref{sec:architecture}, for $ i $-th point cloud scenario $ P_i $ from the source or target domain, LiDAR-based detector generates the BEV features $ \bm{F}_i \in \mathbb{R}^{H \times W \times C}$, where $ H $, $ W $, and $ C $ denote the height, width and channel numbers of the feature map.
We first project the corresponding ground truth $ L^s_i $ or pseudo-labels $ \hat{L}^t_i $ to the BEV feature map, and then randomly extract the equal-length sequences $ \bm{F}_i^{+} \in  \mathbb{R}^{M_i \times C} $ and $\bm{F}_i^{-} \in  \mathbb{R}^{M_i \times C} $.
Here, $ M_i $ is the length of the feature sequence, $ \bm{F}_i^{+} $ and $ \bm{F}_i^{-} $ represent the foreground and background features from BEV, respectively.

\begin{figure}[t]	
	\centering
	\includegraphics[width=\linewidth]{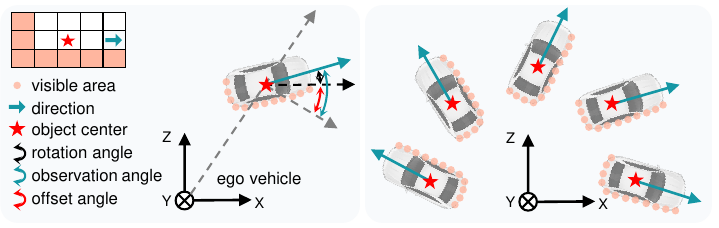}
	%	\vspace{-7mm}	
	\caption{Left: Demonstration of the offset angle.
		Right: Objects with same offset angle share similar geometric structures.}\label{fig_prototype}
	\vspace{-4mm}
\end{figure}

\paragraph{Group.} 
For the extracted foreground features $ \bm{F}_i^{+} $, we further divide them into different groups according to their geometric structures on point clouds.
Specifically, for $ j $-th foreground $ \bm{F}_{i,j}^{+} $ in the sequence ($ j \in [1,  M_i] $), we compute its offset angle $ \theta_{i,j}^{\text{off}} $ as follows:
\vspace{-3mm}
\begin{equation}
	\theta_{i,j}^{\text{off}} = \theta_{i,j}^{\text{obs}} - r_{i,j},
	\vspace{-2mm}
\end{equation}
where $ r_{i,j} $ is the direction, $ \theta_{i,j}^{\text{obs}} $ is the observation angle, as presented in Fig.~\ref{fig_prototype} (left).
Note that the direction $ r_{i,j} $ is provided from the labels $ L^s_i $ and $ \hat{L}^t_i $, while the observation angle $ \theta_{i,j}^{\text{obs}} $ can be computed according to the central position of 3D bounding box. 
Next, all foreground features are split into $ K $ groups, and the group index $ Q_{i,j}$ is formulated as:
\vspace{-2mm}
\begin{equation}\label{eq:prototype}
	Q_{i,j} = \lfloor norm(\theta_{i,j}^{\text{off}})/\delta \rfloor + 1,
	\vspace{-2mm}
\end{equation}
where $ norm(\cdot) $ is a normalization function that converting the input angles into $ [0, 2\pi] $, and $ \delta = 2\pi / K$  is the interval of angles between groups.
In this way, the foreground features with similar offset angles $ \theta_{i,j}^{\text{off}} $ are assigned into the same group, where their geometric structures are very similar, as demonstrated in Fig.~\ref{fig_prototype} (right).
Additionally, the extracted backgrounds $ \bm{F}_{i,j}^{-} $ are sent into an individual group, thus totally $K+1$ groups are built.

\paragraph{Prototype Construction.}
At the beginning of training, we randomly initialize a series of learnable prototypes $ \mathcal{G} = \{ g_k \}_{k=1}^{K+1} \in \mathbb{R}^{(K+1) \times C}$.
During training, we extract the BEV features $ \bm{F}_{i} $ from both source and target domains, and split them into corresponding groups via Eq.~\ref{eq:prototype}.
In $ k $-th group, the foreground features $ \bm{F}_{i,j}^{+} $ are enforced to be aligned with the foreground prototype $ g_{k ( k \in [1,K])}$.
Similarly, the background features $\bm{F}_{i,j}^{-}$ in the last group are aligned with the background prototype $ g_{K+1}$.

\subsection{Soft Contrast Loss}
\label{sec:loss}
Given a point cloud $ P_i $, our goal is to align its fore/background features $ \bm{F}_{i}^{+} $ and $ \bm{F}_{i}^{-} $ with the corresponding prototypes in $ \mathcal{G}$. 

\vspace{-2mm}
\paragraph{Intra-group Attract.} 
For the foreground features $ \bm{F}_{i}^{+} $, we pull them closer with the corresponding prototype in $ \mathcal{G}$, which can be formulated as:
\vspace{-2mm}
\begin{equation}
	\mathcal{L}_{att}^{+} = \sum_{k=1}^{K} \sum_{i=1}^{N} \sum_{j=1}^{M_{i}} (1 - sim(\bm{F}_{i,j}^{+}, \bm{g}_{k}))\mathbbm{1}[Q_{i,j} = k],
	\vspace{-3mm} 
\end{equation}
where $ sim(\bm{a}, \bm{b}) = \frac{\bm{a} \cdot \bm{b}}{\vert\vert \bm{a}\vert\vert \, \vert\vert\bm{b}\vert\vert}$ is the cosine similarity, $ \mathbbm{1}[Q_{i,j} = k] $ is an indicator function that equals to 1 if $ Q_{i,j} = k $ and 0 otherwise.
Similarly, the background features $ \bm{F}_{i}^{-} $ are also required to be pulled to the background prototype $ \bm{g}_{K+1} $, which can be calculated as:
\vspace{-3mm}
\begin{equation}
\mathcal{L}_{att}^{-} = \sum_{i=1}^{N} \sum_{j=1}^{M_{i}} (1 - sim(\bm{F}_{i,j}^{-}, \bm{g}_{K+1})).
\vspace{-5mm}
\end{equation}

\paragraph{Inter-group Repel.} 
To enhance the discriminative capacity, we need to push the features away from all prototypes belonging to other groups.
For example, the distances between background features $  F_i^{-} $ and all foreground prototypes are minimized via:
\vspace{-3mm}
\begin{equation}
	\mathcal{L}_{rep}^{-} = \sum_{k=1}^{K} \sum_{i=1}^{N} \sum_{j=1}^{M_{i}} max(0, sim(\bm{F}_{i,j}^{-}, \bm{g}_{k})).
	\vspace{-3mm}
\end{equation}
For foreground features within adjacent groups, their corresponding geometric structures are relatively more similar.
Repelling these features away is not very necessary, and might even make the training process unstable.
Hence, we adopt a more relaxed constraints as follows:
\vspace{-3mm}
\begin{equation}
	\begin{aligned}
	&\mathcal{L}_{rep}^{+_{adj}} = \sum_{i=1}^{N} \sum_{j=1}^{M_{i}} \sum_{k \in A_{i,j}}  max(0, sim(\bm{F}_{i,j}^{+}, \bm{g}_{k})-m), \\
	&\mathcal{L}_{rep}^{+_{other}} = \sum_{i=1}^{N} \sum_{j=1}^{M_{i}} \sum_{k \notin A_{i,j}, k \neq Q_{i,j}}  max(0, sim(\bm{F}_{i,j}^{+}, \bm{g}_{k})),
	\end{aligned}
\vspace{-2mm}
\end{equation}
where $ m $ indicates the margin which is set to 0.5 in our experiments, $ A_{i,j} $ is the index of the groups adjacent to $ Q_{i,j} $, \textit{i.e.}, $ A_{i,j} = Q_{i,j} \pm 1$.
The overall soft contrast loss $ \mathcal{L}_{contra} $ can be formulated as: 
\vspace{-3mm}
\begin{equation}
	\mathcal{L}_{contra} = \mathcal{L}_{att}^{+} + \mathcal{L}_{att}^{-} + \beta_1\mathcal{L}_{rep}^{+_{adj}} + \beta_2\mathcal{L}_{rep}^{+_{other}} + \beta_3\mathcal{L}_{rep}^{-}, 
	\vspace{-0mm}
\end{equation}
where $ \beta_1 $, $ \beta_2 $ and $ \beta_3 $ are the balance coefficients.

\begin{figure}[t]	
	\centering
	\includegraphics[width=\linewidth]{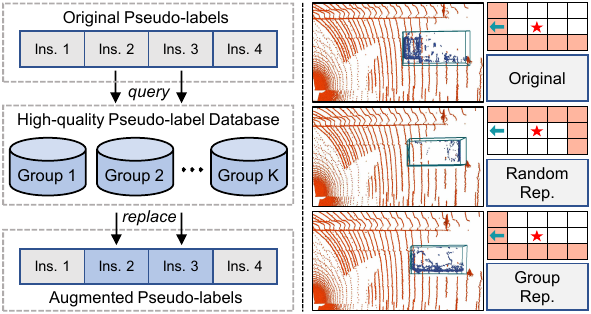}
	%	\vspace{-7mm}	
	\caption{Illustration of instance replacement augmentation (IRA).
		Left: IRA leverages a group mechanism to displace the original instances with high-quality candidates.
		Right: Compared with random replacing, our group mechanism does not interfere the spatial context of the point cloud scene.}\label{fig_insrep}
	\vspace{-4mm}
\end{figure}

\subsection{Noise Sample Suppression}
\label{sec:suppression}

The pseudo-labels used on the target domain are noisy and can lead to the accumulation of errors.
To mitigate the impact of noise, we propose the noise sample suppression (NSS) approach, which generates a noise mask to suppress the magnitude of the gradient descent for the foreground areas that might be underlying noise.
The noise mask can be represented as $ S \in \{\alpha, 1.0 \}^{H\times W} $, where $ \alpha \, (\alpha < 1.0)$  is the suppression factor to decrease the contribution of low-quality samples.
%, thus  
%
In $ S$, the foreground areas that have high similarities with the background prototype, \textit{i.e.}, $ sim(\bm{F}_{i,j}^{+}, \bm{g}_{K+1}) > 0.3 $, are assigned to $ \alpha $, while rest foreground and background areas are assigned to $ 1.0 $.

During training, the noise mask $ S $ is multiplied to the co-training loss $ \mathcal{L}_{co\text{-}train} $, elaborated in Sec.~\ref{sec:overall_training_procedure}. 
With the progress of training, prototypes will be optimized with better representative capability, which enables NSS to suppress the noise more reliably and facilitate the training procedure.

\subsection{Instance Replacement Augmentation}
\label{sec:aug}

\begin{algorithm2e}[t]
	\footnotesize
	\SetKwInOut{Input}{input}\SetKwInOut{Output}{output}
	
	\Input{labeled source domain $ \mathbb{D}^s $, unlabeled target domain $ \mathbb{D}^t $, 3D detector $ \bm{\theta} $, total epochs $ T $, steps per epoch $ N $, and list of update epochs $ U $}
	\Output{adapted 3D detector $ \bm{\theta}^{t} $}
	Pre-train the network $ \bm{\theta}^s  \leftarrow \mathbb{D}^s$ according to Eq.~(\ref{eq:det_loss})\;
	Initialize $ \bm{\theta}  \leftarrow \bm{\theta}^s$\;
	Generate pseudo-labels and database $ (\hat{L}^t, \hat{D}^t) \leftarrow (\bm{\theta}, \mathbb{D}^t)  $\;
	\For{$ epoch\leftarrow 1 $ to $ T $}
	{\For{$ step\leftarrow 1 $ to $ N $}
		{Sample mini-batches $ (\beta^s, \beta^t) \leftarrow (\mathbb{D}^s, \mathbb{D}^t) $\;
			Instance replacement $ \beta^t_{\text{aug}} \leftarrow $ IRA$ (\beta^t, \hat{D}^t) $ \;
			update $ \bm{\theta} \leftarrow (\beta^s, \beta^t_{\text{aug}})$ according to Eq.~(\ref{eq:adapt})\;}
		\If{$ epoch \in U $}{Update pseudo-labels $ \hat{L}^t \leftarrow (\bm{\theta}, \mathbb{D}^t)  $\;}	
	}
	$ \bm{\theta}^{t} \leftarrow \bm{\theta}$\;
	\caption{The learning procedure of GPA-3D}\label{algorithm}
\end{algorithm2e}

Those uncertain pseudo-labels (with scores of 0.2$ \, \sim \, $0.5) are usually ignored in training.
Despite inaccurate, they might provide partial localization information.
To this end, we devise the instance replacement augmentation (IRA) module.
As shown in Fig.~\ref{fig_insrep} (left), we first pick the pseudo-labels with scores over 0.5 to construct a high-quality database, which utilizes the group mechanism as Eq.~\ref{eq:prototype} to divide the picked instances into groups belonging to different geometric structures.
During training, we calculate the group indexes for the uncertain pseudo-labels, and replace them with instances having same group indexes from the database.
In this procedure, a parameter $p_{\textit{IRA}}$ is adopted to regulate the probability of the replacement operation.

There are two main merits of IRA.
First, the quantity of target data is maintained and the diversity is also enhanced.
Second, benefiting from the group mechanism, the spatial contexts around the replaced instances are unchanged and no ambiguous or unreasonable case is introduced, as devised in Fig.~\ref{fig_insrep} (right).

\begin{table*}[t]
	\caption{Comparison with the state-of-the-art methods on the Waymo $ \rightarrow $ KITTI adaptation scenario, with BEV and 3D average precisions of 40 recall positions. 
		In addition, we also report the Closed Gap from ST3D\cite{uda3det_st3d}, which is defined as $ \frac{\text{AP}_\text{model} - \text{AP}_\text{source}}{\text{AP}_\text{oracle} - \text{AP}_\text{source}} \times 100\%$.
		For fair comparison, the results with the detector of SECOND-IoU are obtained from the original paper of ST3D++~\cite{3duda_st3d++}, while the performances with PointPillars are cited from 3D-CoCo~\cite{3duda_3dcoco}.
		The best result is indicated by \textbf{bold}.
	}
	\label{tab_waymo-to-kitti_comparison}
	\vspace{-5mm}
	\begin{center}
		\scriptsize
		\tabcolsep=0.5mm
		\renewcommand{\arraystretch}{0.9}
		\resizebox{\textwidth}{!}{
			\begin{tabular}{p{1.9cm}<{\centering}|p{1.1cm}<{\centering} p{1.3cm}<{\centering} p{1.1cm}<{\centering} p{1.3cm}<{\centering} | p{1.1cm}<{\centering} p{1.3cm}<{\centering} p{1.1cm}<{\centering} p{1.3cm}<{\centering}}
				\toprule
				\multirow{2}{*}{\textbf{Methods}} & \multicolumn{4}{c|}{\textbf{SECOND-IoU}} & \multicolumn{4}{c}{\textbf{PointPillars}} \\
				\cline{2-9}
				\rule{0pt}{0.25cm}
				& \textbf{$ \text{AP}_\text{BEV} $} & \textbf{Closed Gap} & \textbf{$ \text{AP}_\text{3D} $} & \textbf{Closed Gap} & \textbf{$ \text{AP}_\text{BEV} $} &  \textbf{Closed Gap}  & \textbf{$ \text{AP}_\text{3D} $} & \textbf{Closed Gap}\\
				\midrule
				\midrule
				Source Only & 67.64 & - & 27.48 & - & 47.8 & - & 11.5 & - \\
				\midrule
				SN~\cite{uda3det_sn} & 78.96 & $ + $72.33\% & 59.20 & $ + $69.00\% & 27.4& $ - $55.14\% & 6.4& $ - $8.49\%\\
				UMT~\cite{uda3det_uncertainty}  & 77.79 & $ + $64.86\% & 64.56 & $ + $80.66\%& - & - &- & -\\
				3D-CoCo~\cite{3duda_3dcoco}  &  - & - & - & - & 76.1 & $ + $76.49\%& 42.9& $ + $52.25\%\\
				ST3D~\cite{uda3det_st3d} & 82.19 & $ + $92.97\% & 61.83 & $ + $74.72\% & 58.1& $ + $27.84\% & 23.2& $ + $19.47\%\\
				ST3D++~\cite{3duda_st3d++} & 80.78 & $ + $83.96\% & 65.64 & $ + $83.01\% & - & - & - & - \\
				\midrule
				GPA-3D (ours) &\textbf{83.79} & $ + $103.19\% & \textbf{70.88} &$ + $94.41\%& \textbf{77.29}& $ + $79.70\% &\textbf{50.84} & $ + $65.46\%\\
				\rowcolor{MyCyan}\textit{Improvement} & $ + $\textit{1.6} & $ + $\textit{10.22}\% & $ + $\textit{5.24} & $ + $\textit{11.4}\% & $ + $\textit{1.19} & $ + $\textit{3.21}\% & $ + $\textit{7.94} & $ + $\textit{13.21}\% \\
				\midrule
				\midrule
				Oracle &83.29 & - & 73.45 & - & 84.8& - & 71.6& -\\

				\bottomrule
		\end{tabular}}
	\end{center}
	\vspace{-4mm}
\end{table*}

\begin{table*}[t]
	\caption{Adaptation performance on the Waymo $ \rightarrow $ nuScenes  in comparison with different base detectors and state-of-the-art approaches.}
	\label{tab_waymo-to-nuscenes_comparison}
	\vspace{-5mm}
	\begin{center}
		\scriptsize
		\tabcolsep=0.5mm
		\renewcommand{\arraystretch}{0.9}
		\resizebox{\textwidth}{!}{
			\begin{tabular}{p{1.9cm}<{\centering}|p{1.1cm}<{\centering} p{1.3cm}<{\centering} p{1.1cm}<{\centering} p{1.3cm}<{\centering} | p{1.1cm}<{\centering} p{1.3cm}<{\centering} p{1.1cm}<{\centering} p{1.3cm}<{\centering}}
				\toprule
				\multirow{2}{*}{\textbf{Methods}} & \multicolumn{4}{c|}{\textbf{SECOND-IoU}} & \multicolumn{4}{c}{\textbf{PointPillars}} \\
				\cline{2-9}
				\rule{0pt}{0.25cm}
				& \textbf{$ \text{AP}_\text{BEV} $} & \textbf{Closed Gap} & \textbf{$ \text{AP}_\text{3D} $} & \textbf{Closed Gap} & \textbf{$ \text{AP}_\text{BEV} $} &  \textbf{Closed Gap}  & \textbf{$ \text{AP}_\text{3D} $} & \textbf{Closed Gap}\\
				\midrule
				\midrule
				Source Only & 32.91 & - & 17.24 & - & 27.8 & - & 12.1 & - \\
				\midrule
				SN~\cite{uda3det_sn} & 33.23 & $ + $1.69\% & 18.57 & $ + $7.54\% & 28.31& $ + $2.41\% & 12.98& $ + $4.58\%\\
				UMT~\cite{uda3det_uncertainty}  & 35.10 & $ + $11.54\% & 21.05 & $ + $21.61\%& - & - &- & -\\
				3D-CoCo~\cite{3duda_3dcoco}  &  - & - & - & - & 33.1 & $ + $25.00\%& 20.7 & $ + $44.79\%\\
				ST3D~\cite{uda3det_st3d} & 35.92 & $ + $15.87\% & 20.19 & $ + $16.73\% & 30.6& $ + $13.21\% & 15.6& $ + $18.23\%\\
				ST3D++~\cite{3duda_st3d++} & 35.73 & $ + $14.87\% & 20.90 & $ + $20.76\% & - & - & - & - \\
				\midrule
				GPA-3D (ours) & \textbf{37.25} & $ + $22.88\% & \textbf{22.54} & $ + $30.06\% & \textbf{35.47} & $ + $36.18\% & \textbf{21.01}& $ + $46.41\%\\
				\rowcolor{MyCyan}\textit{Improvement} & $ + $\textit{1.33} & $ + $\textit{7.01}\% & $ + $\textit{1.49} & $ + $\textit{8.45}\% & $ + $\textit{2.37} & $ + $\textit{11.18}\% & $ + $\textit{0.31} & $ + $\textit{1.62}\% \\
				\midrule
				Oracle &51.88 & - & 34.87 & - & 49.0& - & 31.3& -\\

				\bottomrule
		\end{tabular}}
	\end{center}
	\vspace{-7mm}
\end{table*}

\subsection{Overall Training Procedure}
\label{sec:overall_training_procedure}
The overall training procedure of GPA-3D is illustrated in Alg.~\ref{algorithm}.
Following previous works~\cite{uda3det_st3d,3duda_st3d++}, the 3D detector is first trained on the labeled source domain $ \mathbb{D}^s $ via minimizing the detection loss $ \mathcal{L}_{det}^{s} $ as:
\vspace{-2mm}
\begin{equation}\label{eq:det_loss}
	\mathcal{L}_{det}^{s} = \mathcal{L}_{reg}^{s} + \mathcal{L}_{cls}^{s},
	\vspace{-2mm}
\end{equation}
where the $ \mathcal{L}_{reg}^{s} $ and $ \mathcal{L}_{cls}^{s} $ indicate the regression and classification errors respectively.
Next, we use the pre-trained detector to generate pseudo-labels $ \hat{L}^t_i $ and the database of IRA on the unlabeled target domain $ \mathbb{D}^t $.
Finally, the co-training paradigm is employed to further fine-tune the model as:
\vspace{-2mm}
\begin{equation}
	\mathcal{L}_{co\text{-}train} = \mathcal{L}_{det}^{s} +  \mathcal{L}_{det}^{t} ,
	\vspace{-2mm}
\end{equation}
where $ \mathcal{L}_{det}^{t} $ is the detection loss on target data, same as in Eq.~\ref{eq:det_loss}.
The overall adaptation loss $ \mathcal{L}_{adapt} $ is calculated via:
\vspace{-4mm}
\begin{equation}\label{eq:adapt}
	\mathcal{L}_{adapt} = \beta \cdot \mathcal{L}_{contra} +  S \cdot \mathcal{L}_{co\text{-}train},
	\vspace{-1mm}
\end{equation}
where $\beta$ is the total weight of the soft contrast loss, and $ S $ is the noise mask of NSS.
For more details of the training procedure, please refer to the supplements.

\section{Experiments}

\subsection{Experimental Setup}

\paragraph{Datasets.}
We evaluate the GPA-3D on widely used autonomous driving benchmarks including Waymo~\cite{dataset_waymo}, nuScenes~\cite{dataset_nuscenes}, and KITTI~\cite{dataset_kitti}. 
These datasets exhibit significant diversities in foreground patterns and LiDAR beams, which can lead to severe domain bias when transferring 3D detectors from one dataset to another.
Detailed information about datasets is available in the supplementary material.

\begin{figure*}[t]	
	\centering
	\includegraphics[width=\textwidth]{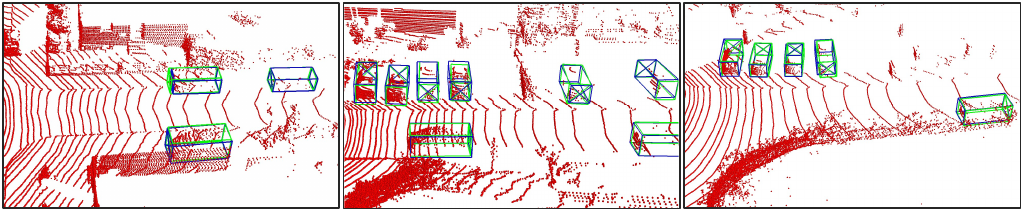}
	%	\vspace{-8mm}
	\caption{Qualitative results of GPA-3D on Waymo $ \rightarrow $ KITTI.
		For each box, we use the X to specify the orientation. 
		The predicted results and ground truths are painted in blue and green, respectively.}\label{fig_vis}
	\vspace{-3mm}
\end{figure*}

\paragraph{Implementation Details.}
We verify the GPA-3D with two popular LiDAR-based detectors, namely SECOND-IoU~\cite{uda3det_st3d} and PointPillars~\cite{pointpillars}. 
All the parameter settings for network architecture are set the same with OpenPCDet~\cite{openpcdet} and ST3D~\cite{uda3det_st3d}. 
We perform all experiments using 8 NVIDIA V100 GPU cards.
For the pre-training step, the model is trained for 30 epochs using the ADAM~\cite{adam} optimizer and the total batch size of 32 on the source domain. 
Next, we utilize the pre-trained model to generate pseudo-labels on the target domain with a score threshold of 0.2. 
Note that instances with scores over 0.5 are retained and subsequently utilized to establish the high-quality pseudo-label database for IRA.
Finally, we further fine-tune the model with our proposed approach for 30 epochs.
To avoid local minima, we employ the cosine annealing strategy to adjust the learning rate, which was set to 0.003 for pre-training and 0.0015 for fine-tuning.
Please refer to the supplements for more implementation details.

\paragraph{Compared Methods.}
As shown in Tab.~\ref{tab_waymo-to-nuscenes_comparison}, GPA-3D is first compared with the Source Only method, which trains the model on the source domain and evaluates it on the target domain without any adaptation.
Next, 5 existing works are included in the comparison, namely, SN~\cite{uda3det_sn}, UMT~\cite{uda3det_uncertainty}, 3D-CoCo~\cite{3duda_3dcoco}, ST3D~\cite{uda3det_st3d}, and ST3D++~\cite{3duda_st3d++}. 
SN utilizes the statistics from target annotations to normalize the foreground objects on the source domain.
UMT employs a mean-teacher framework to filter inaccurate pseudo-labels.
3D-CoCo learns the instance-level transferable features for better generalization. 
ST3D and ST3D++ adopt a memory bank to produce high-quality pseudo-labels.
Additionally, we also compare GPA-3D with the Oracle method, which trains the model on the labeled target data, serving as an upper bound for performance.

\subsection{Comparison with State-of-the-art Methods}

\paragraph{Waymo $ \rightarrow $ KITTI Adaptation.}
To validate the effectiveness to the domain shift about object size, we conduct a comprehensive comparison on Waymo $ \rightarrow $ KITTI.
As demonstrated in Tab.~\ref{tab_waymo-to-kitti_comparison}, with the 3D  detector SECOND-IoU, our proposed GPA-3D outperforms ST3D++~\cite{3duda_st3d++} with a large margin, and significant performance gains are obtained compared with previous best results, \textit{i.e.}, 5.24\% of AP$ _{\text{3D}} $ and 1.6\% of AP$ _{\text{BEV}} $.
Note that the AP$ _{\text{BEV}} $ of GPA-3D is also higher than Oracle method, indicating the effectiveness of incorporating the geometric structure information into UDA on 3D detection task.
Even switching the base detector to PointPillars, our method still exceeds previous SOTA 3D-CoCo~\cite{3duda_3dcoco} by 7.94\% and 1.19\% in terms of AP$ _{\text{3D}} $ and AP$ _{\text{BEV}} $, respectively.

\paragraph{Waymo $ \rightarrow $ nuScenes Adaptation.}
For the domain gap of LiDAR beams, we select Waymo $ \rightarrow $ nuScenes as representatives due to their different LiDAR sensors, \textit{i.e.}, 64-beam \textit{vs} 32-beam.
As shown in Tab.~\ref{tab_waymo-to-nuscenes_comparison}, GPA-3D improves the adaptation performances to 37.25\% AP$ _{\text{BEV}} $ and 22.54\% AP$ _{\text{3D}} $ with the SECOND-IoU detector, surpassing previous SOTA methods.
Compared with ST3D++~\cite{3duda_st3d++}, 1.52\% and 1.64\% gains separately in terms of AP$ _{\text{BEV}}$ and  AP$ _{\text{3D}} $ are achieved.
Based on PointPillars, our approach exceeds the best method 3D-CoCo~\cite{3duda_3dcoco} by 2.37\% in AP$ _{\text{BEV}}$, and outperforms ST3D~\cite{uda3det_st3d} with 4.87\% and 5.41\% respectively in terms of AP$ _{\text{BEV}}$ and AP$ _{\text{3D}}$.
These improvements demonstrate the advancement of our GPA-3D to mitigate the more challenging domain shift of cross-beam scenarios.

\begin{table}[t]
	\caption{Component ablation studies in GPA-3D. 
		\textbf{Proto} indicates the geometry-aware prototype alignment.
		\textbf{Soft} is the soft contrast loss.
		\textbf{NSS} means the noise sample filtering.
		\textbf{IRA} represents the instance replacement augmentation.}
	%	\vspace{-3.5mm}
	\label{tab_ablation_components}
	\centering
	\scriptsize
	\renewcommand{\arraystretch}{1.0}
	\resizebox{\linewidth}{!}{
		\begin{tabular}{c | c c c c | c c}
			\toprule
			\textbf{Setting} & \textbf{Proto} & \textbf{Soft} & \textbf{NSS} & \textbf{IRA} & \textbf{$\text{AP}_\text{BEV}$} & \textbf{$\text{AP}_\text{3D}$} \\
			\midrule
			\rowcolor{MyGray}(a) & & & & & 77.87 & 60.36 \\
			(b) & $ \checkmark $ &  & & & 80.49  & 66.28\\
			(c) & $ \checkmark $ & $ \checkmark $ & & & 80.51 & 67.34\\
			(d) & $ \checkmark $ & $ \checkmark $ & $ \checkmark $ & & 83.07  & 69.45\\
			(e) & $ \checkmark $ & $ \checkmark $ & & $ \checkmark $ & 81.94 & 67.79 \\
			\rowcolor{MyCyan}(f) & $ \checkmark $ & $ \checkmark $ & $ \checkmark $ & $ \checkmark $ & \textbf{83.79} & \textbf{70.88}\\
			\bottomrule	
		\end{tabular}}
%					
%					1 & d& & & & & 77.87 / 60.36 
%									
			
	\vspace{-5mm}
\end{table}

\subsection{Ablation Studies}

All ablation studies are conducted on Waymo $ \rightarrow $ KITTI with SECOND-IoU as the base detector.

\paragraph{Component Analysis in GPA-3D.}
We assess the effectiveness of each component in GPA-3D, as presented in Tab.~\ref{tab_ablation_components}.
Baseline (a) represents self-training via pseudo-labels on the target domain.
The application of geometry-aware prototype alignment provides 5.92\% and 2.62\% gains separately in terms of AP$ _{\text{3D}} $ and AP$ _{\text{BEV}} $, and the soft contrast loss brings an improvement of 1.06\% on AP$ _{\text{3D}} $.
The improvements demonstrate that incorporating the geometric relationship into domain adaptation is feasible and effective.
In addition, NSS and IRA boost the performance by around 2.5\% and 1.5\% respectively, which indicates
the efficacy of enhancing the quality of supervision on target data.

\begin{figure}[t]	
	\centering
	\includegraphics[width=\linewidth]{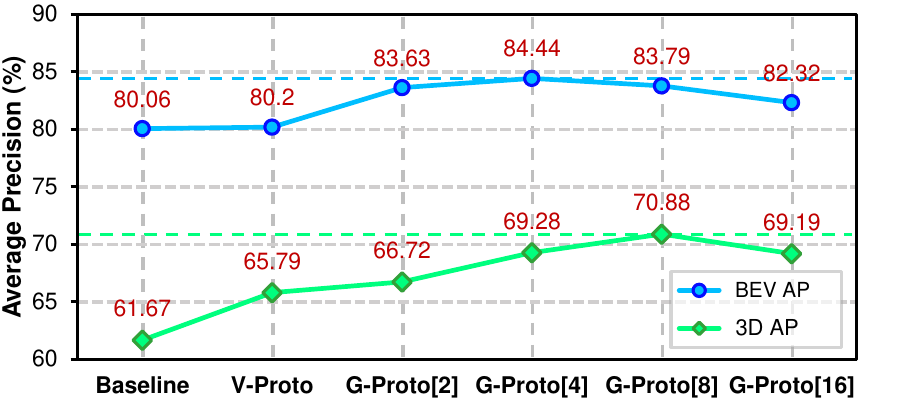}
	%	\vspace{-7mm}	
	\caption{Ablations on the geometry-aware prototype alignment. 
		\textbf{Baseline} is the co-training method without any feature alignment.
		\textbf{V-Proto} refers to the vanilla alignment with a pair of fore/background prototypes.
		\textbf{G-Proto[$ \bm{n} $]} indicates that $ n $ prototypes are employed in GPA-3D. 
	}\label{fig_ab_proto}
	\vspace{-4mm}
\end{figure}

\paragraph{Effectiveness of Geometry-aware Prototype Alignment.}
We further investigate the effects of the geometry-aware prototype alignment.
As illustrated in Fig.~\ref{fig_ab_proto}, the vanilla alignment with one pair of fore/background prototypes performs better than the co-training baseline, implying that the misalignment of features distribution affects the performance.
Applying two prototypes yields 3.57\% and 5.05\% gains of AP$ _{\text{BEV}}$ and AP$ _{\text{3D}}$ respectively, compared to the co-training baseline.
The performance reaches to the peak of 84.44\% AP$ _{\text{BEV}}$ when 4 foreground prototypes are employed, indicating the advancement of combining geometric information with feature alignment.
However, we observe minor performance degradation when too many prototypes are used, which we attribute to redundant prototypes leading to indistinguishable features in the representational space.

\begin{table}[t]
	\caption{Ablations on noise sample suppression.
		The symbols -T/-S denote that NSS is applied solely on the target/source domain, while -TS performs NSS on both target and source domains. -TSH additionally adopts a hard truncated factor, \textit{i.e.}, $ \alpha=0 $.}
	%	\vspace{-3.5mm}
	\label{tab_ablation_mask}
	\centering
	\scriptsize
	\renewcommand{\arraystretch}{1.0}
	\resizebox{\linewidth}{!}{
		\begin{tabular}{l |p{1.7cm}<{\centering}  p{0.8cm}<{\centering}| p{1.0cm}<{\centering}  p{1.0cm}<{\centering}}
			\toprule
			\textbf{Methods} & \textbf{Filter Domain}
			& $ \bm{\alpha} $  & \textbf{$ \text{AP}_\text{BEV}$} & \textbf{$ \text{AP}_\text{3D}$} \\	
			\midrule
			\rowcolor{MyGray}GPA-3D (w/o NSS) &  -  & - & 81.94 & 67.79\\
			\midrule
			GPA-3D (w/ NSS-T) &  Target  & 0.5 & 83.37 & 68.24 \\
			GPA-3D (w/ NSS-S) &  Source  & 0.5 & 82.33 & 67.93 \\
			GPA-3D (w/ NSS-TS) &  Target + Source  & 0.5 & 83.45 & 69.77\\
			\rowcolor{MyCyan}GPA-3D (w/ NSS-TSH) &  Target + Source  & 0.0 & 83.79 & 70.88\\
			
			\bottomrule
	\end{tabular}}
	\vspace{-1mm}
\end{table}

\paragraph{Effectiveness of Noise Sample Suppression.}
We conduct ablations on the noise sample filter (NSS) with various settings.
As shown in Tab.~\ref{tab_ablation_mask}, the detection performance drops to 67.79\% AP$ _{\text{3D}}$, when we remove the NSS from GPA-3D.
Only applying the NSS on target domain achieves the gains of 1.43\% and 0.45\% on AP$ _{\text{BEV}}$ and AP$ _{\text{3D}}$, respectively.
We could see that using NSS on the source domain could also bring improvements.
We think this is due to the fact that NSS suppresses those source samples with only a few points, which are very similar to the background noise.
When the hard truncated $ \alpha $ is adopted, AP$ _{\text{3D}}$
is further improved to 70.88\%, indicating the effectiveness of NSS.

\begin{table}[t]
	\caption{Effects of the instance replacement augmentation. 
		\textbf{RandRep} discards the group mechanism in IRA.}
	%	\vspace{-3.5mm}
	\label{tab:ab_ira}
	\centering
	\scriptsize
	\renewcommand{\arraystretch}{1.0}
	\tabcolsep=1mm
	\resizebox{\linewidth}{!}{
		\begin{tabular}{p{1.9cm}<{\centering} | >{\columncolor{MyGray}}c  p{1.9cm}<{\centering} >{\columncolor{MyCyan}}c}
			\toprule
			\textbf{Method} & \textbf{w/o IRA} & \textbf{RandRep} & \textbf{w/ IRA} \\
			\midrule
			\textbf{$\text{AP}_\text{BEV}$} / \textbf{$\text{AP}_\text{3D}$}& 83.07 / 69.45 & 82.99 / 69.59 & 83.79 / 70.88\\
			\bottomrule	
	\end{tabular}}

	\vspace{-1mm}
\end{table}

\paragraph{Effectiveness of Instance Replacement Augmentation.}
Also, we compare different policies in instance replacement augmentation (IRA).
We can see from Tab.~\ref{tab:ab_ira} that our proposed IRA attains 0.72\% and 1.43\% gains in terms of $\text{AP}_\text{BEV}$ and $\text{AP}_\text{3D}$, respectively.
Without the group mechanism in IRA, \textit{i.e.}, randomly replacing pseudo-labels with instances the database, only marginal gains are obtained in $\text{AP}_\text{3D}$, and even degradation in $\text{AP}_\text{BEV}$.
This highlights the significance of maintaining the consistency between instances and their contextual environments.

\begin{table}[t]
	\caption{Comparison with different adaptation frameworks. 
		\textbf{Source} refers to the Source Only method.
		\textbf{Self-T.} is the self-training framework.
		\textbf{Co-T.} symbolizes the co-training pipeline.
		\textbf{Mean T.} represents the mean teacher paradigm.}
	%	\vspace{-3.5mm}
	\label{tab:ablation_framework}
	\centering
	\scriptsize
	\renewcommand{\arraystretch}{1.0}
	\tabcolsep=1mm
	\resizebox{\linewidth}{!}{
		\begin{tabular}{c | >{\columncolor{MyGray}}c  c c c >{\columncolor{MyCyan}}c}
			\toprule
			\textbf{Framework} & \textbf{Source} & \textbf{Self-T.} & \textbf{Co-T.} & \textbf{Mean T.}   & \textbf{GPA-3D} \\
			\midrule
			\textbf{$\text{AP}_\text{BEV}$} / \textbf{$\text{AP}_\text{3D}$}& 67.64 / 27.48 & 77.87 / 60.36 & 80.06 / 61.67 & 80.01 / 64.62   & \textbf{83.79} / \textbf{70.88}\\
			\bottomrule	
	\end{tabular}}
	%					
	%					1 & d& & & & & 77.87 / 60.36 
	%									
	
	%	\vspace{-6mm}
\end{table}

\begin{figure}[t]	
	\centering
	\includegraphics[width=\linewidth]{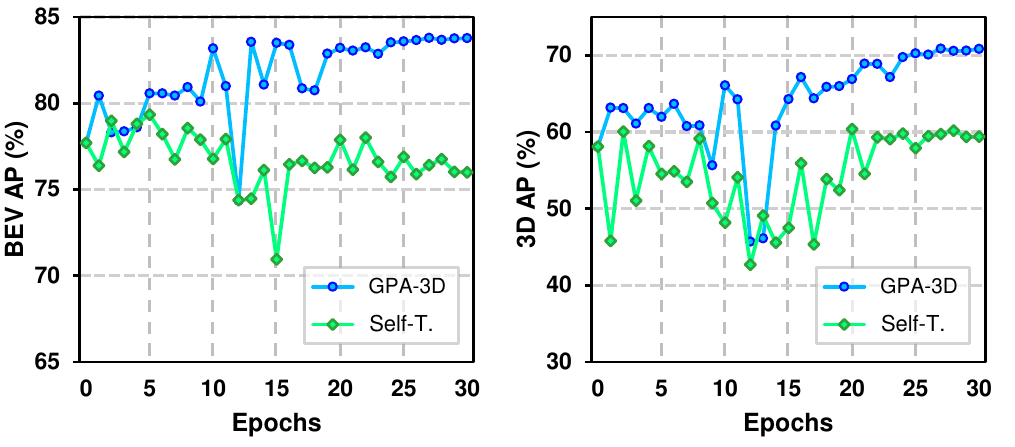}
	\vspace{-5mm}	
	\caption{Comparisons of self-training baseline and our GPA-3D.}\label{fig_result_comparison}
	\vspace{-3mm}
\end{figure}

\paragraph{Domain Adaptation Frameworks.}
We compare our proposed GPA-3D with several adaptation frameworks, as presented in Tab.~\ref{tab:ablation_framework}.
The results confirm the effectiveness of GPA-3D, which leverages the geometric association to transfer 3D detectors across different domains.
Fig.~\ref{fig_result_comparison} further illustrates that, despite all models fluctuate at early epochs, our GPA-3D steadily and consistently enhances the detection performance in later training stages.

\begin{figure}[t]	
	\centering
	\subfigure[Source only]{
		\includegraphics[width=0.49\linewidth]{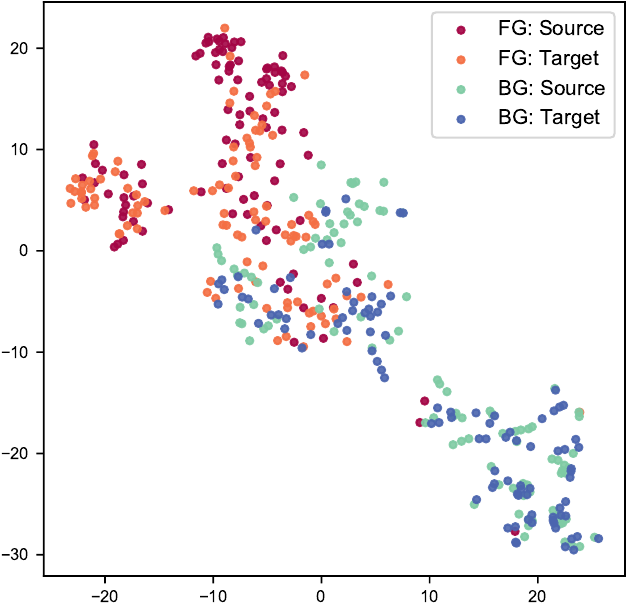}}\hspace{-1mm}
	\subfigure[Our method]{
		\includegraphics[width=0.49\linewidth]{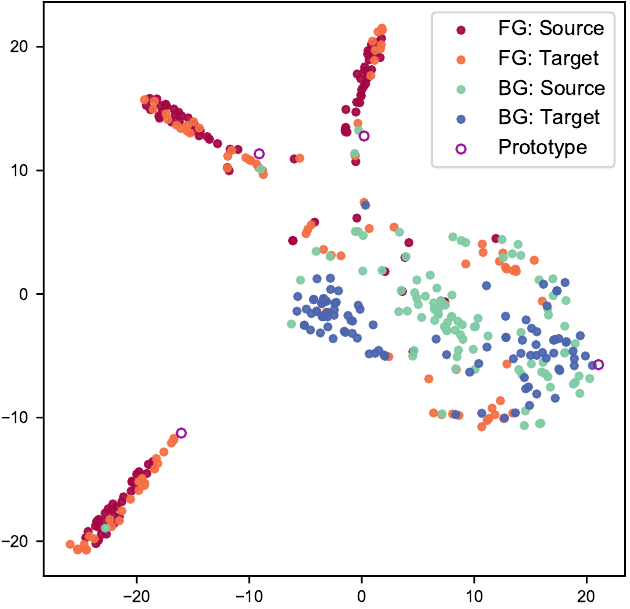}}
	%	\vspace{-7mm}	
	\caption{The t-SNE visualization of BEV features.
	The feature points are obtained by SECOND-IoU on Waymo $ \rightarrow $ nuScenes.}\label{fig_tsne}
	\vspace{-4mm}
\end{figure}

\vspace{-4mm}
\paragraph{Visualization.}
We exhibit some qualitative results of cross-domain adaptation in Fig.~\ref{fig_vis}. 
Additionally, in Fig.~\ref{fig_tsne}, we visualize the distribution of BEV features.
It is obvious that GPA-3D aggregates foreground samples into different prototypes, and separates them from the backgrounds.
Further visualizations can be found in in the supplements.

\section{Conclusion}
This paper presents a novel framework for unsupervised domain adaptive 3D detection.
Our proposed GPA-3D leverages the underlying geometric relationship to reduce the distributional discrepancy in the feature space, thus mitigating the domain shift problems.
Comprehensive experiments demonstrate that our method is effective and can be easily incorporated into mainstream LiDAR-based 3D detectors.
For future work, we plan to extend GPA-3D to support multi-modal 3D detectors.
This requires a more efficient alignment mechanism to process feature streams from both point clouds and images.

\section*{Acknowledgement}
This work was supported by the National Natural Science Foundation of China under No.62276061 and 62006041.

{\small
	\bibliographystyle{ieee_fullname}
	\bibliography{ref}

\begin{thebibliography}{10}\itemsep=-1pt

\bibitem{survey_arnold2019survey}
Eduardo Arnold, Omar~Y Al-Jarrah, Mehrdad Dianati, Saber Fallah, David Oxtoby,
  and Alex Mouzakitis.
\newblock A survey on {3D} object detection methods for autonomous driving
  applications.
\newblock {\em TITS}, 20(10):3782--3795, 2019.

\bibitem{dataset_nuscenes}
Holger Caesar, Varun Bankiti, Alex~H Lang, Sourabh Vora, Venice~Erin Liong,
  Qiang Xu, Anush Krishnan, Yu Pan, Giancarlo Baldan, and Oscar Beijbom.
\newblock nuscenes: A multimodal dataset for autonomous driving.
\newblock In {\em CVPR}, pages 11621--11631, 2020.

\bibitem{dafasterrcnn}
Yuhua Chen, Wen Li, Christos Sakaridis, Dengxin Dai, and Luc Van~Gool.
\newblock Domain adaptive faster r-cnn for object detection in the wild.
\newblock In {\em CVPR}, pages 3339--3348, 2018.

\bibitem{3ddet_rangedet}
Lue Fan, Xuan Xiong, Feng Wang, Naiyan Wang, and Zhaoxiang Zhang.
\newblock Rangedet: In defense of range view for lidar-based {3D} object
  detection.
\newblock In {\em ICCV}, pages 2918--2927, 2021.

\bibitem{domainadversarial}
Yaroslav Ganin, Evgeniya Ustinova, Hana Ajakan, Pascal Germain, Hugo
  Larochelle, Fran{\c{c}}ois Laviolette, Mario Marchand, and Victor Lempitsky.
\newblock Domain-adversarial training of neural networks.
\newblock {\em JMLR}, 17(1):2096--2030, 2016.

\bibitem{dataset_kitti}
Andreas Geiger, Philip Lenz, and Raquel Urtasun.
\newblock Are we ready for autonomous driving? the kitti vision benchmark
  suite.
\newblock In {\em CVPR}, pages 3354--3361, 2012.

\bibitem{survey_guo2020deep}
Yulan Guo, Hanyun Wang, Qingyong Hu, Hao Liu, Li Liu, and Mohammed Bennamoun.
\newblock Deep learning for {3D} point clouds: A survey.
\newblock {\em TPAMI}, 43(12):4338--4364, 2020.

\bibitem{survey_3dpointclouds}
Yulan Guo, Hanyun Wang, Qingyong Hu, Hao Liu, Li Liu, and Mohammed Bennamoun.
\newblock Deep learning for {3D} point clouds: A survey.
\newblock {\em TPAMI}, 2020.

\bibitem{uda3det_uncertainty}
Deepti Hegde, Vishwanath Sindagi, Velat Kilic, A~Brinton Cooper, Mark Foster,
  and Vishal Patel.
\newblock Uncertainty-aware mean teacher for source-free unsupervised domain
  adaptive {3D} object detection.
\newblock {\em arXiv preprint arXiv:2109.14651}, 2021.

\bibitem{hsu2020progressive}
Han-Kai Hsu, Chun-Han Yao, Yi-Hsuan Tsai, Wei-Chih Hung, Hung-Yu Tseng, Maneesh
  Singh, and Ming-Hsuan Yang.
\newblock Progressive domain adaptation for object detection.
\newblock In {\em WCCV}, pages 749--757, 2020.

\bibitem{huang2018auggan}
Sheng-Wei Huang, Che-Tsung Lin, Shu-Ping Chen, Yen-Yi Wu, Po-Hao Hsu, and
  Shang-Hong Lai.
\newblock Auggan: Cross domain adaptation with gan-based data augmentation.
\newblock In {\em ECCV}, pages 718--731, 2018.

\bibitem{2d_uda_seg_proca}
Zhengkai Jiang, Yuxi Li, Ceyuan Yang, Peng Gao, Yabiao Wang, Ying Tai, and
  Chengjie Wang.
\newblock Prototypical contrast adaptation for domain adaptive semantic
  segmentation.
\newblock In {\em ECCV}, pages 36--54. Springer, 2022.

\bibitem{khodabandeh2019robust}
Mehran Khodabandeh, Arash Vahdat, Mani Ranjbar, and William~G Macready.
\newblock A robust learning approach to domain adaptive object detection.
\newblock In {\em ICCV}, pages 480--490, 2019.

\bibitem{kim2019self}
Seunghyeon Kim, Jaehoon Choi, Taekyung Kim, and Changick Kim.
\newblock Self-training and adversarial background regularization for
  unsupervised domain adaptive one-stage object detection.
\newblock In {\em ICCV}, pages 6092--6101, 2019.

\bibitem{adam}
Diederik~P Kingma and Jimmy Ba.
\newblock Adam: A method for stochastic optimization.
\newblock {\em arXiv preprint arXiv:1412.6980}, 2014.

\bibitem{pointpillars}
Alex~H Lang, Sourabh Vora, Holger Caesar, Lubing Zhou, Jiong Yang, and Oscar
  Beijbom.
\newblock Pointpillars: Fast encoders for object detection from point clouds.
\newblock In {\em CVPR}, pages 12697--12705, 2019.

\bibitem{li2022stepwise}
Guofa Li, Zefeng Ji, and Xingda Qu.
\newblock Stepwise domain adaptation (sda) for object detection in autonomous
  vehicles using an adaptive centernet.
\newblock {\em TITS}, 2022.

\bibitem{survey_lidarpointclouds}
Ying Li, Lingfei Ma, Zilong Zhong, Fei Liu, Michael~A Chapman, Dongpu Cao, and
  Jonathan Li.
\newblock Deep learning for lidar point clouds in autonomous driving: a review.
\newblock {\em TNLLS}, 2020.

\bibitem{2d_uda_prototype_continual}
Hongbin Lin, Yifan Zhang, Zhen Qiu, Shuaicheng Niu, Chuang Gan, Yanxia Liu, and
  Mingkui Tan.
\newblock Prototype-guided continual adaptation for class-incremental
  unsupervised domain adaptation.
\newblock In {\em ECCV}, pages 351--368. Springer, 2022.

\bibitem{3duda_mlcnet}
Zhipeng Luo, Zhongang Cai, Changqing Zhou, Gongjie Zhang, Haiyu Zhao, Shuai Yi,
  Shijian Lu, Hongsheng Li, Shanghang Zhang, and Ziwei Liu.
\newblock Unsupervised domain adaptive {3D} detection with multi-level
  consistency.
\newblock In {\em ICCV}, pages 8866--8875, 2021.

\bibitem{dataset_once}
Jiageng Mao, Minzhe Niu, Chenhan Jiang, Jingheng Chen, Xiaodan Liang, Yamin Li,
  Chaoqiang Ye, Wei Zhang, Zhenguo Li, Jie Yu, et~al.
\newblock One million scenes for autonomous driving: Once dataset.
\newblock In {\em NIPS}, 2021.

\bibitem{survey_mao20223d}
Jiageng Mao, Shaoshuai Shi, Xiaogang Wang, and Hongsheng Li.
\newblock {3D} object detection for autonomous driving: a review and new
  outlooks.
\newblock {\em arXiv preprint arXiv:2206.09474}, 2022.

\bibitem{pointnet}
Charles~R Qi, Hao Su, Kaichun Mo, and Leonidas~J Guibas.
\newblock Pointnet: Deep learning on point sets for {3D} classification and
  segmentation.
\newblock In {\em CVPR}, pages 652--660, 2017.

\bibitem{pointnet++}
Charles~Ruizhongtai Qi, Li Yi, Hao Su, and Leonidas~J Guibas.
\newblock Pointnet++: Deep hierarchical feature learning on point sets in a
  metric space.
\newblock In {\em NIPS}, pages 5099--5108, 2017.

\bibitem{rodriguez2019domain}
Adrian~Lopez Rodriguez and Krystian Mikolajczyk.
\newblock Domain adaptation for object detection via style consistency.
\newblock {\em arXiv preprint arXiv:1911.10033}, 2019.

\bibitem{strong_weak}
Kuniaki Saito, Yoshitaka Ushiku, Tatsuya Harada, and Kate Saenko.
\newblock Strong-weak distribution alignment for adaptive object detection.
\newblock In {\em CVPR}, pages 6956--6965, 2019.

\bibitem{uda3det_sfuda3d}
Cristiano Saltori, St{\'e}phane Lathuili{\`e}re, Nicu Sebe, Elisa Ricci, and
  Fabio Galasso.
\newblock Sf-uda {3D}: Source-free unsupervised domain adaptation for
  lidar-based {3D} object detection.
\newblock In {\em 3DV}, pages 771--780, 2020.

\bibitem{pvrcnn}
Shaoshuai Shi, Chaoxu Guo, Li Jiang, Zhe Wang, Jianping Shi, Xiaogang Wang, and
  Hongsheng Li.
\newblock Pv-rcnn: Point-voxel feature set abstraction for {3D} object
  detection.
\newblock In {\em CVPR}, pages 10529--10538, 2020.

\bibitem{pointrcnn}
Shaoshuai Shi, Xiaogang Wang, and Hongsheng Li.
\newblock Pointrcnn: {3D} object proposal generation and detection from point
  cloud.
\newblock In {\em CVPR}, pages 770--779, 2019.

\bibitem{dataset_waymo}
Pei Sun, Henrik Kretzschmar, Xerxes Dotiwalla, Aurelien Chouard, Vijaysai
  Patnaik, Paul Tsui, James Guo, Yin Zhou, Yuning Chai, Benjamin Caine, et~al.
\newblock Scalability in perception for autonomous driving: Waymo open dataset.
\newblock In {\em CVPR}, pages 2446--2454, 2020.

\bibitem{3ddet_rsn}
Pei Sun, Weiyue Wang, Yuning Chai, Gamaleldin Elsayed, Alex Bewley, Xiao Zhang,
  Cristian Sminchisescu, and Dragomir Anguelov.
\newblock Rsn: Range sparse net for efficient, accurate lidar {3D} object
  detection.
\newblock In {\em CVPR}, pages 5725--5734, 2021.

\bibitem{2d_uda_prototype_oriented}
Korawat Tanwisuth, Xinjie Fan, Huangjie Zheng, Shujian Zhang, Hao Zhang, Bo
  Chen, and Mingyuan Zhou.
\newblock A prototype-oriented framework for unsupervised domain adaptation.
\newblock {\em Advances in Neural Information Processing Systems},
  34:17194--17208, 2021.

\bibitem{openpcdet}
OpenPCDet~Development Team.
\newblock Openpcdet: An open-source toolbox for {3D} object detection from
  point clouds.
\newblock \url{https://github.com/open-mmlab/OpenPCDet}, 2020.

\bibitem{uda3det_sn}
Yan Wang, Xiangyu Chen, Yurong You, Li~Erran Li, Bharath Hariharan, Mark
  Campbell, Kilian~Q Weinberger, and Wei-Lun Chao.
\newblock Train in germany, test in the usa: Making {3D} object detectors
  generalize.
\newblock In {\em CVPR}, pages 11713--11723, 2020.

\bibitem{second}
Yan Yan, Yuxing Mao, and Bo Li.
\newblock Second: Sparsely embedded convolutional detection.
\newblock {\em Sensors}, 18(10):3337, 2018.

\bibitem{3ddet_pixor}
Bin Yang, Wenjie Luo, and Raquel Urtasun.
\newblock Pixor: Real-time {3D} object detection from point clouds.
\newblock In {\em CVPR}, pages 7652--7660, 2018.

\bibitem{uda3det_st3d}
Jihan Yang, Shaoshuai Shi, Zhe Wang, Hongsheng Li, and Xiaojuan Qi.
\newblock St3d: Self-training for unsupervised domain adaptation on {3D} object
  detection.
\newblock In {\em CVPR}, pages 10368--10378, 2021.

\bibitem{3duda_st3d++}
Jihan Yang, Shaoshuai Shi, Zhe Wang, Hongsheng Li, and Xiaojuan Qi.
\newblock St3d++: Denoised self-training for unsupervised domain adaptation on
  {3D} object detection.
\newblock {\em TPAMI}, 2022.

\bibitem{3dssd}
Zetong Yang, Yanan Sun, Shu Liu, and Jiaya Jia.
\newblock 3dssd: Point-based {3D} single stage object detector.
\newblock In {\em CVPR}, pages 11040--11048, 2020.

\bibitem{3duda_3dcoco}
Zeng Yihan, Chunwei Wang, Yunbo Wang, Hang Xu, Chaoqiang Ye, Zhen Yang, and
  Chao Ma.
\newblock Learning transferable features for point cloud detection via {3D}
  contrastive co-training.
\newblock {\em NIPS}, 34:21493--21504, 2021.

\bibitem{2d_uda_detection_mttrans}
Jinze Yu, Jiaming Liu, Xiaobao Wei, Haoyi Zhou, Yohei Nakata, Denis Gudovskiy,
  Tomoyuki Okuno, Jianxin Li, Kurt Keutzer, and Shanghang Zhang.
\newblock Cross-domain object detection with mean-teacher transformer.
\newblock In {\em ECCV}, 2022.

\bibitem{3ddet_iassd}
Yifan Zhang, Qingyong Hu, Guoquan Xu, Yanxin Ma, Jianwei Wan, and Yulan Guo.
\newblock Not all points are equal: Learning highly efficient point-based
  detectors for {3D} lidar point clouds.
\newblock In {\em CVPR}, pages 18953--18962, 2022.

\bibitem{cyclegan}
Jun-Yan Zhu, Taesung Park, Phillip Isola, and Alexei~A Efros.
\newblock Unpaired image-to-image translation using cycle-consistent
  adversarial networks.
\newblock In {\em ICCV}, pages 2223--2232, 2017.

\end{thebibliography}
}

\appendix
\section{Overview}
\label{sec:supp_overview}
This document presents additional technical details, and provides both quantitative and qualitative results to support the submitted paper. 
In Sec.~\ref{sec:supp_datasets}, we discuss the large-scale datasets
used in the experiments, and analyze their intrinsic characteristics that cause severe domain shifts. 
In Sec.~\ref{sec:supp_more_implementation}, we elaborate on
the network architectures of the 3D detectors employed for comparisons, and describe the implementation details of
GPA-3D. 
In Sec.~\ref{sec:supp_more_vis}, we offer more comprehensive quantitative results and visualizations of our approach.

\section{Datasets}
\label{sec:supp_datasets}
\begin{figure}[t]	
	\centering
	\includegraphics[width=\linewidth]{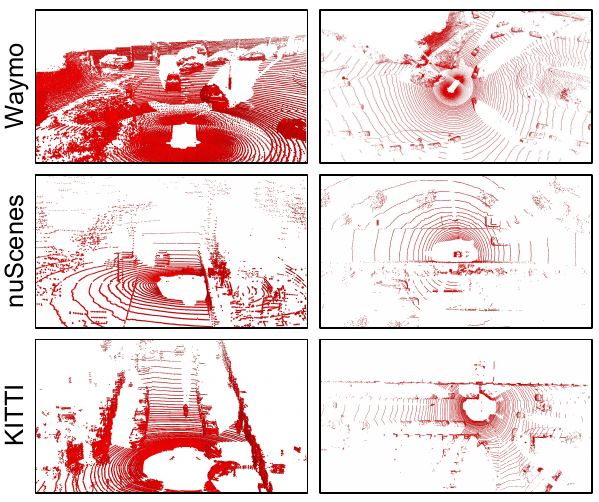}
	\vspace{-1mm}	
	\caption{Visualizations of the point clouds for different datasets. 
		Left: Frontal view.
		Right: Bird's-eye view.}\label{fig:dataset}
	\vspace{-1mm}
\end{figure}

We conduct comprehensive experiments on the prevalent autonomous driving datasets, namely Waymo~\cite{dataset_waymo}, nuScenes~\cite{dataset_nuscenes}, and KITTI~\cite{dataset_kitti}. 
These datasets have diverse weather conditions, sensor configurations, foreground styles, and annotation quantities, thereby causing serious domain shifts when adapting a LiDAR-based 3D detector from one dataset to another. 
Fig.~\ref{fig:dataset} presents randomly selected examples from the aforementioned datasets. 
Subsequently, we will introduce each dataset in detail.

\paragraph{Waymo.}
For recent 3D detection
task, Waymo~\cite{dataset_waymo} is the
most large-scale and challenge benchmark, which includes 798 sequences (more than 150,000 frames) for training and 202 sequences (approximately 40,000 frames)
for validation.
Waymo provides the point clouds captured by a 64-beam LiDAR and 4 200-beam blind LiDAR for each frame.
In our experiments, we use the 1.2 version of Waymo and subsample only 50\% of the training samples, consistent with ST3D~\cite{uda3det_st3d} and ST3D++~\cite{3duda_st3d++}.

\paragraph{nuScenes.}
The nuScenes~\cite{dataset_nuscenes} dataset comprises of 28,130 samples in the training set and 6,019 samples in the validation set. 
Point clouds within nuScenes are captured by a 32-beam LiDAR in Boston and Singapore, under diverse weather conditions. 
To ensure consistency with previous works, we access the performance of transferring 3D detectors across different LiDAR beams by treating all 28,130 training scenes as the target domain.

\paragraph{KITTI.}
As a popular autonomous driving dataset, KITTI~\cite{dataset_kitti} contains 7,481
labeled frames for training and 7,518 unlabeled frames for testing. 
The point clouds of KITTI are captured by a 64-beam Velodyne LiDAR in Karlsruhe, Germany. 
Following previous approaches, we partition the training frames into two distinct sets: the \textit{train} split, comprising 3,712 samples, and the \textit{val} split, consisting of 3,769 samples.

\section{More Implementation Details}
\label{sec:supp_more_implementation}

\paragraph{Co-training Framework.}
We follow the default settings of ONCE~\cite{dataset_once}, an open-source 3D detection codebase, to construct the co-training framework in GPA-3D. 
Specifically, this co-training framework feeds an equal number of point clouds from both source and target domains into the 3D detector in each mini-batch.
The outputs generated by the detector are then used for loss computation, with the supervision of ground truth and pseudo-labels, respectively. 
The calculated losses are subsequently summed together to update the detector parameters and prototypes via the back-propagation method.

\paragraph{Detection Architecture.}
To ensure fair comparisons, we adopt the default configurations of ST3D~\cite{uda3det_st3d} and  ST3D++~\cite{3duda_st3d++} to set the voxel size in SECOND-IoU~\cite{second} and PointPillars~\cite{pointpillars} to (0.1$ m $, 0.1$ m $, 0.15$ m $) and (0.2$ m $, 0.2$ m $), respectively.
Furthermore, for all datasets utilized in our experiments, we shift the coordinate origins to the ground plane, and separately set the detection ranges of $ X $, $ Y $, $ Z $ axes to [-75.2$ m $, 75.2$ m $], [-75.2$ m $, 75.2$ m $], and [-2$ m $, 4$ m $].

\paragraph{Hyper-parameters in GPA-3D.}
For the geometry-aware prototype alignment, we set the length $ M_i $ of the feature sequences to be equal to the number of foreground areas in the $ i $-th BEV feature map.
Additionally, we set the prototype numbers to 8 and 4 for the adaptation scenarios of Waymo $ \rightarrow $ KITTI and Waymo $ \rightarrow $ nuScenes, respectively.
For the soft contrast loss, we determine the balance coefficients $ \beta_1 $, $ \beta_2 $, and $ \beta_3 $ to be 5, 1, and 5, respectively.
In our implementation, we perform the instance replacement augmentation with the probability $p_{\textit{IRA}}$ of 0.25.

\section{Exploration Studies}
\label{sec:supp_more_vis}

\paragraph{Extend GPA-3D to Multiple Categories.}
For autonomous driving vehicles, the detection of pedestrians on the road is also a crucial aspect. 
In fact, it is easy and effective to extend GPA-3D to other classes.
Compared to cars, the geometric variations of pedestrians are smaller, thus we reduce the prototype numbers to 3 for pedestrian.
As shown in Tab.~\ref{tab:pedestrian_detection}, GPA-3D improves the pedestrian detection performances to 48.17\% $\text{AP}_\text{BEV}$ and 45.20\% $\text{AP}_\text{3D}$, surpassing previous state-of-the-art methods.
Compared to ST3D++~\cite{3duda_st3d++}, our approach achieves 0.97\% and 1.3\% gains in terms of $\text{AP}_\text{BEV}$ and $\text{AP}_\text{3D}$, respectively.
These improvements demonstrate that GPA-3D has consistent effectiveness on the pedestrian detection.

\begin{table}[t]
	\caption{Comparison with previous works on the pedestrian category. 
		The adaptation scenario is nuScenes $ \rightarrow $ KITTI, and the base detector is SECOND-IoU~\cite{second}. 
		For fair comparison, the results are cited from the original paper of ST3D++~\cite{3duda_st3d++}.
	}
	%	\vspace{-3.5mm}
	\label{tab:pedestrian_detection}
	\centering
	\scriptsize
	\renewcommand{\arraystretch}{1.0}
	\tabcolsep=1mm
	\resizebox{\linewidth}{!}{
		\begin{tabular}{p{1.5cm}<{\centering} | p{1.1cm}<{\centering}  p{1.3cm}<{\centering} p{1.1cm}<{\centering} p{1.3cm}<{\centering}}
			\toprule
			\textbf{Method} & \textbf{$\text{AP}_\text{BEV}$} & \textbf{Closed Gap} & \textbf{$\text{AP}_\text{3D}$} &  \textbf{Closed Gap} \\
			\midrule
			\midrule
			Source Only & 39.95  & -&  34.57 & -\\
			SN~\cite{uda3det_sn}  & 38.91  & $ - $16.07\%&  34.36 & $ - $3.11\%\\
			ST3D~\cite{uda3det_st3d} & 44.00  & $ + $60.36\% &  42.60 & $ + $118.79\% \\
			ST3D++~\cite{3duda_st3d++} & 47.20  &$ + $108.41\% &  43.96 & $ + $138.91\%\\
			\midrule
			GPA-3D (ours) & \textbf{48.17}  & \textbf{$ + $122.97\%}&  \textbf{45.20} & \textbf{$ + $157.25\%}\\
			\rowcolor{MyCyan}\textit{Improvement} & $ + $\textit{0.97}  & $ + $\textit{14.56}\%& $ + $\textit{1.3}  & $ + $\textit{18.34}\%\\
			\midrule
			\midrule
			Oracle & 46.64  & -&  41.33 & -\\
			\bottomrule	
	\end{tabular}}
	%					
	%					1 & d& & & & & 77.87 / 60.36 
	%									
	
	\vspace{-1mm}
\end{table}

\paragraph{Why Could Adaptation Method Outperforms the Oracle.}
In the adaptation scenario of Waymo $\rightarrow$ KITTI, the $\text{AP}_\text{BEV}$ of GPA-3D has surpassed that of the Oracle method, which is fully supervised by the ground truth of KITTI dataset.
We attribute the reason into two aspects.
\textit{1) Label-insufficient target domain:} 
Compared to Waymo, KITTI is a relatively label-insufficient dataset (7,000 \textit{vs.} 150,000).
The limited annotations affect the performance of Oracle.
\textit{2) Stronger generalization ability:} 
Our method reduces the feature discrepancy across domains, bringing stronger generalization ability.
This makes it easier for model to apply the knowledge learned from source domain to the target domain, thereby improving the final performance.

\begin{table}[t]
	\caption{Analysis of different alignment schemes in GPA-3D on Waymo $ \rightarrow $ nuScenes.
		\textbf{Conv.} indicates that an extra branch with three convolution layers are attached to the BEV features for alignment.
		\textbf{Pre.} means to align the intermediate features from the backbone network.
		\textbf{BEV} is the BEV-level alignment in GPA-3D.
	}
	\vspace{1mm}
	\label{tab:align_policy}
	\centering
	\scriptsize
	\renewcommand{\arraystretch}{1.0}
	\tabcolsep=1mm
	\resizebox{\linewidth}{!}{
		\begin{tabular}{p{1.5cm}<{\centering} | p{1.4cm}<{\centering}  p{1.4cm}<{\centering} p{1.4cm}<{\centering} p{1.4cm}<{\centering}}
			\toprule
			\textbf{Method} & \textbf{w/o align} & \textbf{Conv.} & \textbf{Pre.} &  \textbf{BEV} \\
			\midrule
			\textbf{$\text{AP}_\text{BEV}$} / \textbf{$\text{AP}_\text{3D}$} & 35.34 / 20.13 & 35.92 / 22.37 & 35.72 / 22.13  & \textbf{37.25} / \textbf{22.54} \\
			\bottomrule	
	\end{tabular}}
	
	\vspace{1mm}
\end{table}

\begin{table}[t]
	\vspace{1mm}
	\caption{Comparison on nuScenes $\rightarrow$ KITTI with PointRCNN~\cite{pointrcnn}.}
	\label{tab_pointrcnn}
	\vspace{-6mm}
	\begin{center}
		\setlength\tabcolsep{4pt}
		\renewcommand{\arraystretch}{0.9}
		\resizebox{\linewidth}{!}{
			\begin{tabular}{l m{1.6cm}<{\centering} m{1.6cm}<{\centering} m{1.6cm}<{\centering} m{1.6cm}<{\centering} >{\columncolor{MyCyan}}m{1.6cm}<{\centering}}
				\toprule
				Method & SF-UDA$^{\text{3D}}$ & Dreaming & MLC-Net & ST3D++ & GPA-3D \\
				Reference & [3DV'20] & [ICRA'22] & [ICCV'21] & [TPAMI'22] & (ours)\\
				\midrule
				0.7 IoU $\text{AP}_\text{3D}$  & 54.5 & -  & 55.42  & 67.51  &   \textbf{67.77}	\\	
				0.5 IoU $\text{AP}_\text{3D}$  &  - & 70.3 &  - &  79.93 &   \textbf{81.06}	\\		
				\bottomrule				
		\end{tabular}}
	\end{center}
	\vspace{-8mm}
\end{table}

\paragraph{Analysis of Different Alignment Schemes.}
We investigate the effects of different alignment schemes in GPA-3D, as shown in Tab.~\ref{tab:align_policy}.
Without alignment, the adaptation performance degrades due to the distributional discrepancy in the feature space.
Compared with the policies of Conv. and Pre., our BEV-level alignment achieves superior results, indicating the effectiveness of our approach in directly dealing with the distributional discrepancy problem at BEV features.

\paragraph{Extend GPA-3D to Point-based Architecture.}
We also try to extend GPA-3D to a point-based 3D detector, PointRCNN~\cite{pointrcnn}.
For the point-wise features, we assign prototypes to them based on the geometric information of the objects to which they belong.
The results on nuScens$\rightarrow$KITTI demonstrate that GPA-3D has the potential to be applied to point-based detectors with minor adjustments.

\begin{figure*}[t]	
	\centering
	\includegraphics[width=\textwidth]{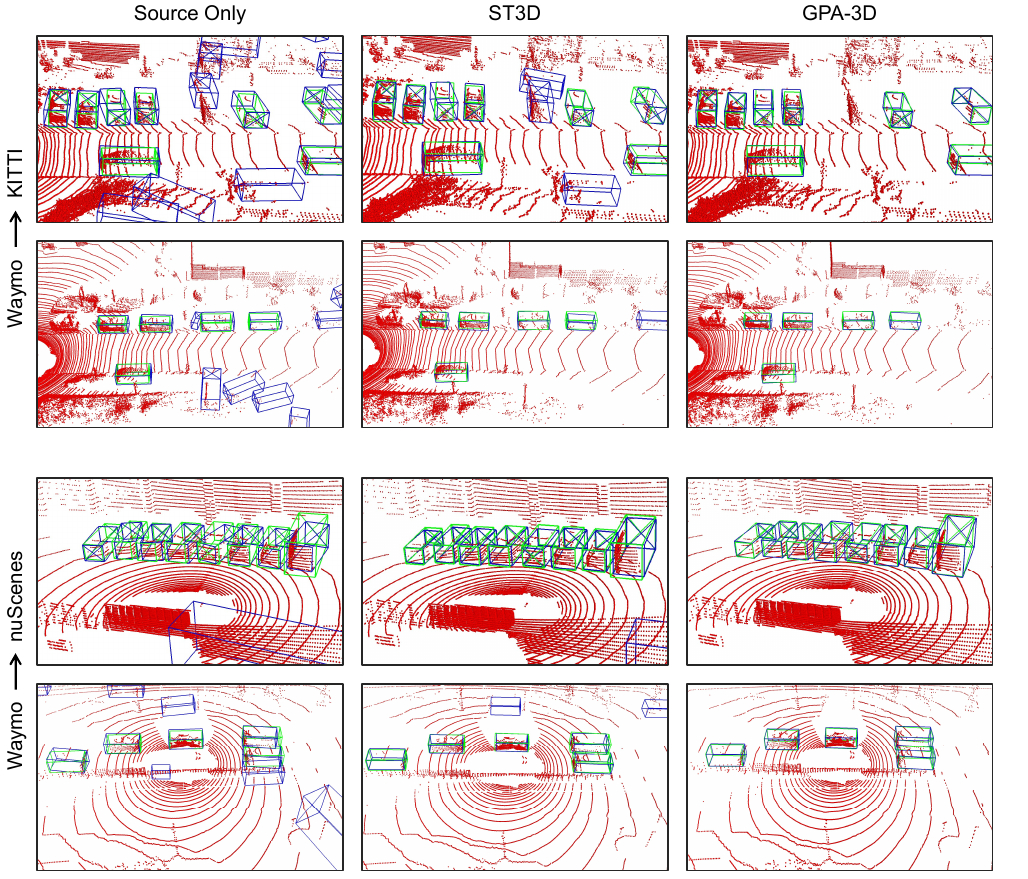}
	%	\vspace{-8mm}
	\caption{Qualitative results on the adaptation scenarios of Waymo $ \rightarrow $ KITTI and Waymo $ \rightarrow $ nuScenes.
		For each box, we use the X to specify the orientation. 
		The predicted results and ground truths are painted in blue and green, respectively.}\label{fig_vis_compare}
	\vspace{-0mm}
\end{figure*}

\begin{figure*}[t]	
	\centering
	\includegraphics[width=\textwidth]{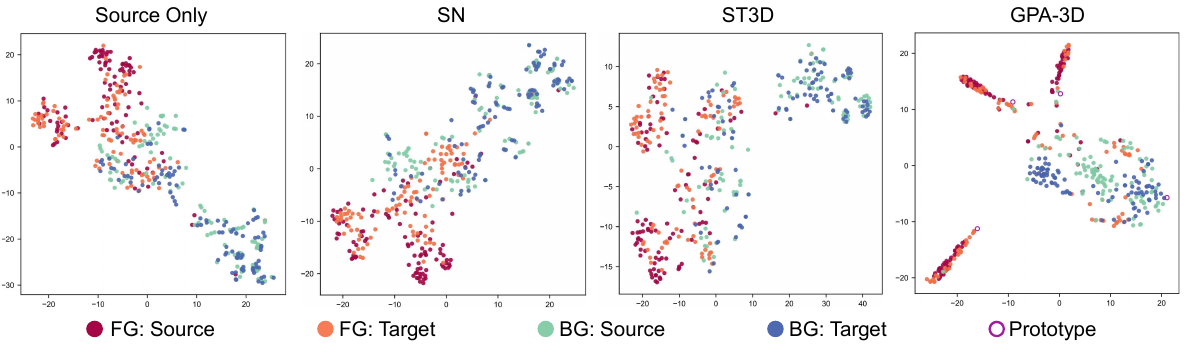}
	%	\vspace{-8mm}
	\caption{The t-SNE visualization of different methods on Waymo $ \rightarrow $ nuScenes.
		SECOND-IoU~\cite{second} is adopted as the base detector.
	}\label{fig_tsne_compare}
	\vspace{-4mm}
\end{figure*}

\paragraph{Qualitative Results.}
We present more visualizations on the adaptation scenarios of Waymo $ \rightarrow $ KITTI and Waymo $ \rightarrow $ nuScenes in Fig.~\ref{fig_vis_compare}.
These qualitative results demonstrate the effectiveness of GPA-3D in improving adaptation performance via reducing the false positive predictions and enhancing the regression accuracy.
To further validate the efficacy of our GPA-3D, we employ the t-SNE method to visualize the feature distributions of different approaches, as illustrated in Fig.~\ref{fig_tsne_compare}.
The results clearly show that GPA-3D clusters the features of the same category in different domains, while also separates the features of different categories.
This indicates that GPA-3D provides better alignment of features and facilitates the transferring across domains.

\end{document}